\documentclass[a4paper,11pt]{article}

\usepackage{geometry}
\usepackage[utf8]{inputenc}
\usepackage{amsmath, amssymb, amsfonts}
\usepackage{graphicx}
\usepackage{booktabs} 
\usepackage[colorlinks=true, linkcolor=blue, citecolor=blue, urlcolor=blue]{hyperref} 
\usepackage{caption}
\usepackage{subcaption}
\usepackage{float}
\usepackage{placeins} 
\usepackage{authblk} 
\usepackage[table]{xcolor}
\usepackage{multirow}
\usepackage{bookmark} 
\usepackage{eso-pic}
\usepackage{rotating}
\usepackage{mathptmx} 

\usepackage{booktabs}
\usepackage{tabularx}
\usepackage{array}
\newcolumntype{Y}{>{\raggedright\arraybackslash}X} 

\title{\textbf{SurgMotion: A Video-Native Foundation Model for Universal Understanding of Surgical Videos}}
\author[1,6]{Jinlin Wu}
\author[3]{Felix Holm}
\author[1]{Chuxi Chen}
\author[4]{An Wang}
\author[1]{Yaxin Hu}
\author[7]{Xiaofan Ye}
\author[1]{Zelin Zang\linebreak}
\author[1,5,6]{Miao Xu}
\author[1]{Lihua Zhou}
\author[11]{Huarui Wu}
\author[8]{Huai Liao}
\author[9]{Danny T. M. CHAN}
\author[10]{Ming Feng\linebreak}
\author[7]{Wai S. Poon}
\author[4]{Hongliang Ren}
\author[1]{Dong Yi}
\author[3]{Nassir Navab}
\author[1,5,6]{Gaofeng Meng}
\author[2]{Jiebo Luo}
\author[1,6]{Hongbin Liu}
\author[1,5,6]{Zhen Lei\thanks{Corresponding author.}}

\affil[1]{Center for Artificial Intelligence and Robotics, Hong Kong Institute of Science and Innovation, Chinese Academy of Sciences, Hong Kong, China}
\affil[2]{Hong Kong Institute of Science and Innovation, Chinese Academy of Sciences, Hong Kong, China}
\affil[3]{Computer Aided Medical Procedures, Technical University of Munich, Munich, Germany}
\affil[4]{Electronic Engineering Department, The Chinese University of Hong Kong, Hong Kong, China}
\affil[5]{University of Chinese Academy of Sciences, Beijing, China}
\affil[6]{State Key Laboratory of Multimodal Artificial Intelligence Systems, Institute of Automation, Chinese Academy of Sciences, Beijing, China}
\affil[7]{Neuromedical Centre, Hong Kong University Shenzhen Hospital, Shenzhen, China}
\affil[8]{Department of Respiratory Medicine, The First Affiliated Hospital of Sun Yat-sen University, Guangzhou, China}
\affil[9]{Department of Surgery, The Chinese University of Hong Kong, Hong Kong, China}
\affil[10]{Department of Neurosurgery, China Pituitary Disease Registry Center, Peking Union Medical College Hospital, Chinese Academy of Medical Sciences and Peking Union Medical College, Beijing, China}
\affil[11]{United Imaging Surgical Co., Ltd.}

\date{\today}

\begin{document} 

\maketitle
\enlargethispage{2\baselineskip} 


    \begin{abstract}
While foundation models have advanced surgical video analysis, current approaches rely predominantly on pixel-level reconstruction objectives that waste model capacity on low-level visual details—such as smoke, specular reflections, and fluid motion—rather than semantic structures essential for surgical understanding. We present SurgMotion, a video-native foundation model that shifts the learning paradigm from pixel-level reconstruction to latent motion prediction. Built on the Video Joint Embedding Predictive Architecture (V-JEPA), SurgMotion introduces three key technical innovations tailored to surgical videos: 1) motion-guided latent masked prediction to prioritize semantically meaningful regions, 2) spatiotemporal affinity self-distillation to enforce relational consistency, and 3) spatiotemporal feature diversity regularization (SFDR) to prevent representation collapse in texture-sparse surgical scenes. To enable large-scale pretraining, we curate SurgMotion-15M, the largest surgical video dataset to date, comprising 3,658 hours of video from 50 sources across 13 organs. Extensive experiments across 17 benchmarks demonstrate that SurgMotion significantly outperforms state-of-the-art methods on surgical workflow recognition (+14.6\% F1 on EgoSurgery, +10.3\% on PitVis), action triplet recognition (39.54\% mAP-IVT on CholecT50), skill assessment, polyp segmentation, and depth estimation. These results establish SurgMotion as a new standard for universal, motion-oriented surgical video understanding. Code and project resources are available at
\href{https://github.com/CAIR-HKISI/SurgMotion}{GitHub}
and
\href{https://surgmotion.cares-copilot.com/}{the project website}.
\end{abstract}


\section{Introduction}

\begin{figure}[!htbp]
    \centering
    \includegraphics[width=0.9\textwidth]{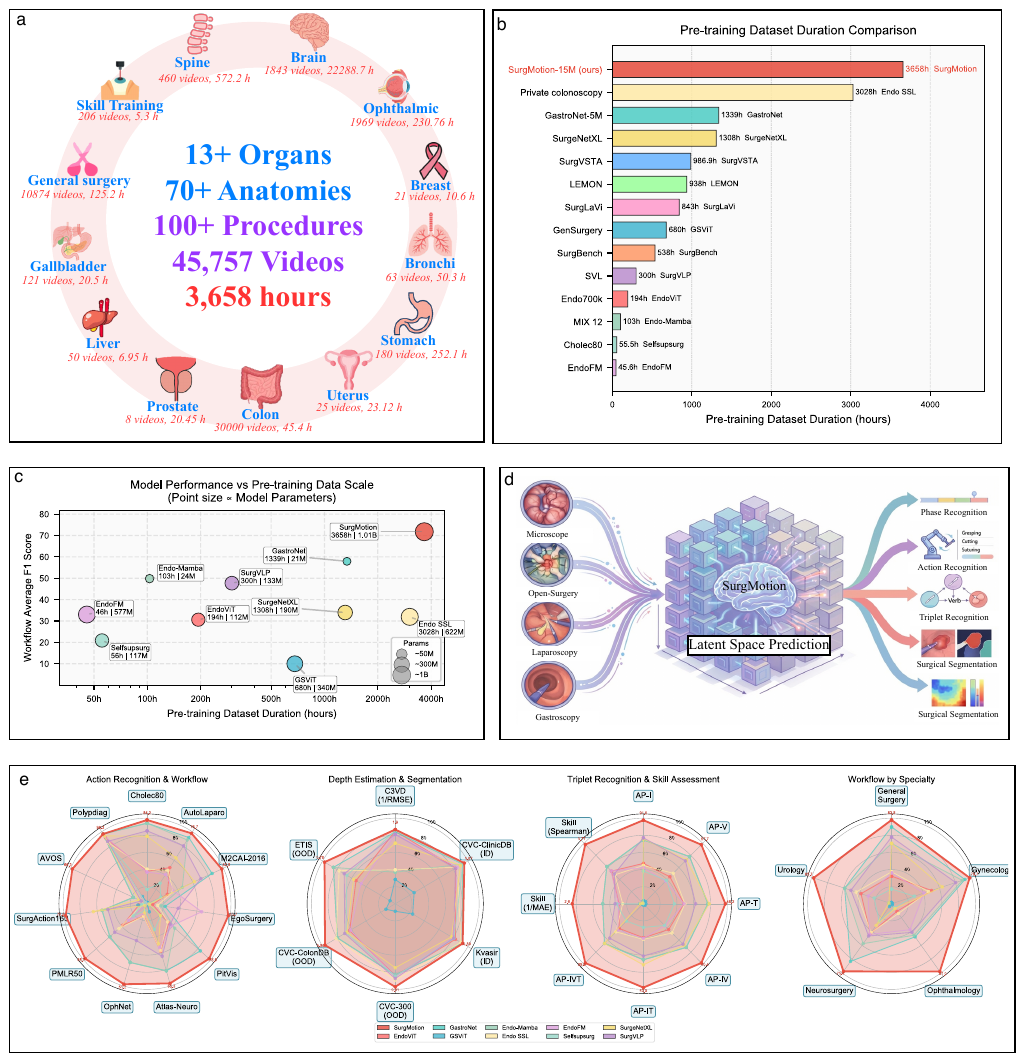}
\caption{\textbf{Overview of SurgMotion.} 
(a) SurgMotion-15M is a large-scale surgical video dataset with 3,658 hours from 50 sources, spanning 12+ organs, 70+ anatomical regions, and 100+ procedures. 
(b) SurgMotion greatly expands the scale of surgical pre-training data. 
(c) Larger pre-training data leads to better workflow recognition performance; marker size denotes model size. 
(d) SurgMotion learns universal representations through motion-guided latent prediction, spatiotemporal self-distillation, and feature diversity regularization. 
(e) The learned representations transfer effectively across diverse downstream tasks and specialties.}
    \label{fig:teaser}
\end{figure}


The development of foundation models has transformed computer vision, with self-supervised pretraining enabling powerful representations that generalize across diverse tasks~\cite{oquab2023dinov2, he2022masked, feichtenhofer2022masked}. In surgical AI, this paradigm shift has led to models such as EndoViT~\cite{batic2024endovit}, EndoFM~\cite{wang2023endofm}, and GSViT~\cite{schmidgall2024general}, which leverage large-scale pretraining for improved performance on downstream tasks. However, despite these advances, fundamental gaps persist that limit the effectiveness of current surgical foundation models and prevent them from achieving robust, universal surgical understanding.

First, current self-supervised pretraining approaches face distinct technical challenges. Current approaches rely predominantly on pixel-level reconstruction objectives borrowed from general vision models such as MAE~\cite{he2022masked} or VideoMAE~\cite{tong2022videomae, wang2023videomae}. In surgical scenes, this creates a fundamental misalignment: models spend capacity learning to reconstruct high-frequency, non-semantic details instead of focusing on the critical yet sparse semantics, such as motion cues of instrument-tissue interactions. 
Furthermore, surgical videos exhibit strong temporal continuity and structured relationships between anatomical regions and tools. However, many pretraining pipelines lack explicit constraints to enforce relational stability across time~\cite{ramesh2023dissecting}. This absence often leads to feature drift and reduced robustness when the model faces significant motion or viewpoint changes, which are common in real surgical procedures.
Compounding these challenges are texture-sparse and visually homogeneous regions, such as exposed adipose tissue or smooth organ surfaces. In self-supervised learning, this increases the risk of low-variance features and channel redundancy. Preventing representation collapse while maintaining diverse, informative feature channels in these monotonous scenarios remains a significant challenge.

Second, beyond architectural limitations, current surgical video models are predominantly confined to small-scale, single-procedure datasets. Common benchmarks focus only on a single type of procedure, such as Cholec80~\cite{twinanda2016endonet} for laparoscopic cholecystectomy or PitVis~\cite{das2025pitvis} for endonasal pituitary surgery. This narrow exposure prevents models from learning universal representations capable of handling the diverse variability of real-world clinical environments. Consequently, existing models struggle to generalize across different anatomical regions, surgical modalities, and institutions~\cite{fujii2024egosurgery}, failing to leverage the potential benefits of scaling up data diversity. Taken together, these challenges call for a video-native surgical foundation model that (i) emphasizes motion-centric semantics beyond pixel reconstruction, (ii) enforces stable spatiotemporal relations, (iii) maintains feature diversity in texture-sparse scenes, and (iv) scales to substantially broader data.
Figure~\ref{fig:teaser} provides an overview of our solution: the scale and diversity of SurgMotion-15M (a--c), the proposed pretraining design (d), and the resulting improvements across tasks and specialties (e).
\FloatBarrier

To address these fundamental limitations, we present \textbf{SurgMotion}, a video-native foundation model designed for universal surgical understanding.
Guided by the overview in Figure~\ref{fig:teaser}(d), our approach introduces three key technical innovations and one dataset contribution that directly target the identified gaps: \begin{itemize}
    \item \textbf{Motion-Guided Latent Masked Prediction.} To address the mismatch between objectives and semantics, we shift masked prediction from pixel space to latent space, building upon the Video Joint Embedding Predictive Architecture (V-JEPA)~\cite{bardes2024revisiting, assran2025vjepa2}. We introduce a motion saliency signal based on spatiotemporal gradients to reweight the prediction loss. This strategy effectively reduces sensitivity to low-level visual disturbances and prioritizes motion-informative regions, ensuring that the model focuses on critical tool dynamics and tissue deformations. 
    \item \textbf{Spatiotemporal Consistency via Affinity Self-Distillation.} To enforce relational stability and provide useful coherence constraints, we introduce a Global Token Affinity Matrix that models the pairwise relationships between all latent tokens. We employ a self-distillation mechanism that aligns the affinity structures of the student network with a momentum-updated teacher via exponential moving average (EMA)~\cite{grill2020bootstrap}, explicitly imposing spatiotemporal coherence and robustness against augmentations without requiring manual labels.
    \item  \textbf{Spatiotemporal Feature Diversity Regularization (SFDR).} To mitigate representation collapse in homogeneous scenes, we introduce SFDR, which operates on two levels: (i) regularizing the variance of pairwise feature similarities to maintain diversity, and (ii) minimizing inter-channel covariance to reduce redundancy. This encourages the network to learn complementary and informative feature channels even in texture-sparse surgical environments.
    \item  \textbf{The SurgMotion-15M dataset.} To overcome the generalization limitations of narrow datasets, we compile SurgMotion-15M, a massive dataset designed to facilitate large-scale self-supervised learning. Aggregating 3,658 hours of video from 50 distinct sources and 13 anatomical regions, this corpus exposes the model to unprecedented variability. This extensive pretraining enables the model to learn universal representations, yielding superior performance across a wide range of downstream validation benchmarks. 
\end{itemize}

Extensive experiments demonstrate that SurgMotion significantly outperforms existing surgical foundation models across a comprehensive range of tasks. On surgical workflow recognition, SurgMotion improves F1-scores by +14.6\% on EgoSurgery and +10.3\% on PitVis compared to the best prior methods. On action triplet recognition, SurgMotion achieves 39.54\% mAP-IVT, setting a new state-of-the-art on CholecT50. On dense prediction tasks, SurgMotion attains superior polyp segmentation under domain shift and best-in-class depth estimation on C3VD. These results establish SurgMotion as a new standard for universal, motion-aware surgical scene understanding. A consolidated comparison across benchmarks and specialties is summarized in Figure~\ref{fig:teaser}(e).

\section{Related Work}

\paragraph{General Vision Foundation Models.}
Foundation models have transformed computer vision by shifting the paradigm from supervised learning on curated datasets to large-scale self-supervised pretraining on unlabeled data~\cite{oquab2023dinov2, he2022masked}. The dominant approach has been masked modeling, where portions of the input are masked and reconstructed by the model. In the image domain, MAE~\cite{he2022masked} demonstrated that Vision Transformers can learn strong representations by reconstructing masked patches. This principle was extended to video through VideoMAE~\cite{tong2022videomae} and VideoMAE V2~\cite{wang2023videomae}, which reconstruct masked spatiotemporal cubes, achieving impressive results on action recognition benchmarks. More recently, the InternVideo family~\cite{wang2024internvideo2,wang2025internvideonext} combines masked video modeling with cross-modal contrastive learning and scales to billions of parameters, achieving state-of-the-art results on general video understanding tasks.

However, pixel-level reconstruction has inherent limitations: models are forced to predict high-frequency visual details that may be stochastic, irrelevant, or fundamentally unpredictable from the available context. This is particularly problematic in domains with high visual noise, such as surgical video, where smoke, specular reflections, and fluid motion introduce unpredictable pixel variations. To address this, Joint Embedding Predictive Architectures (JEPAs) propose predictions in latent space rather than pixel space~\cite{bardes2024revisiting, assran2025vjepa2}. A key challenge with latent prediction is representational collapse, where the model learns trivial solutions. Effective strategies to prevent collapse include teacher-student architectures with exponential moving average (EMA) updates, as employed in DINO~\cite{caron2021emerging}, DINOv2~\cite{oquab2023dinov2}, DINOv3~\cite{seitzer2025dinov3}, BYOL~\cite{grill2020bootstrap}, and V-JEPA~\cite{bardes2024revisiting, assran2025vjepa2}. V-JEPA is particularly relevant for video understanding, as it learns to predict the latent representations of masked spatiotemporal regions, encouraging the model to capture abstract semantic structures and long-range temporal dynamics rather than low-level pixel statistics.

\paragraph{Surgical Foundation Models.}
The success of foundation models in general vision has motivated significant efforts to develop domain-specific models for surgery. Early works focused on static frame analysis. EndoViT~\cite{batic2024endovit} applies masked image modeling to the Endo700k dataset, demonstrating strong transfer to segmentation tasks. GastroNet~\cite{jong2025gastronet} scales image-based pretraining to 5 million gastrointestinal endoscopy images, achieving robustness across different imaging devices and institutions. However, these image-based approaches fundamentally ignore the temporal dynamics that are central to understanding surgical procedures.

To capture procedural dynamics, several video-based surgical foundation models have emerged. EndoFM~\cite{wang2023endofm} pretrains a Video Transformer on 33,000 endoscopic clips using masked reconstruction with spatial-temporal attention mechanisms. EndoSSL~\cite{hirsch2023self} explores Masked Siamese Networks for self-supervised video analysis in colonoscopy and laparoscopy. More recently, EndoMamba~\cite{tian2025endomamba} leverages State Space Models for efficient hierarchical pretraining, enabling real-time video understanding. GSViT~\cite{schmidgall2024general} introduces a generative approach using forward video prediction (next-frame reconstruction) for general surgery. SurgeNetXL~\cite{jaspers2025scaling} scales DINO-based self-supervised learning to 4.7 million frames across multiple procedures, demonstrating the benefits of increased data diversity. Beyond pure vision, SurgVLP~\cite{yuan2025surgvlp} integrates language supervision by aligning surgical video clips with transcribed text from lecture videos, enabling zero-shot recognition capabilities. The SelfSupSurg benchmark~\cite{ramesh2023dissecting} provides a systematic comparison of self-supervised methods (MoCo v2, SimCLR, SwAV, DINO) specifically for surgical computer vision, establishing important baselines for the field.

Despite this progress, existing surgical foundation models face persistent limitations. Most rely on pixel-level reconstruction objectives, which are inefficient for the high-noise surgical environment where models waste capacity on unpredictable visual artifacts such as smoke, specular reflections, and fluid motion. \textbf{SurgMotion} addresses these limitations by building upon the V-JEPA framework, adopting its paradigm of predicting in latent space rather than pixel space, while introducing surgical-specific innovations including motion-guided masking, spatiotemporal consistency constraints, and feature diversity regularization tailored to the unique challenges of surgical video.

\paragraph{Surgical Video Datasets.}
The development of surgical AI has been fundamentally constrained by data availability. Unlike natural image datasets that can be scraped from the internet, surgical data requires institutional partnerships, ethics approval, and careful de-identification to address privacy regulations. Furthermore, meaningful annotations require domain expertise from surgeons or trained medical personnel, making large-scale labeling prohibitively expensive.

Existing surgical datasets are typically isolated efforts targeting specific procedures and tasks. For laparoscopic surgery, Cholec80~\cite{twinanda2016endonet}, M2CAI~\cite{stauder2017tumlap}, and CholecT50~\cite{nwoye2022rendezvous} focus on cholecystectomy, while AutoLaparo~\cite{wang2022autolaparo} and PmLR50~\cite{guo2025pmlr50} cover hysterectomy and liver resection respectively. Other procedures are represented by PitVis~\cite{das2025pitvis} for pituitary surgery and OphNet~\cite{hu2024ophnet} for ophthalmic procedures. Open surgery datasets include EgoSurgery~\cite{fujii2024egosurgery}, AVOS~\cite{goodman2024avos}, and skill assessment benchmarks such as JIGSAWS~\cite{ahmidi2017dataset} and AIxSuture~\cite{hoffmann2024aixsuture}. Additional benchmarks include SurgicalActions160~\cite{schoeffmann2018surgicalactions} for laparoscopic action recognition and PolypDiag~\cite{tian2022contrastive} for polyp detection in colonoscopy. Colonoscopy datasets~\cite{bernal2015cvc, rau2019colonoscopydepth, bobrow2023c3vd} provide annotations for polyp segmentation, depth estimation, and 3D reconstruction.

While individually valuable, these datasets create fragmented ``data silos'' that prevent models from learning universal surgical concepts. Each dataset covers a single site, procedure type, and annotation schema, making aggregation non-trivial. Our \textbf{SurgMotion-15M} dataset addresses this fragmentation by systematically aggregating 3,658 hours of video from 50 distinct sources spanning 13 anatomical regions, providing the scale and diversity necessary for a truly generalizable surgical foundation model.

\section{Methodology}
\label{sec:method}

\begin{table*}[t]
\centering
\small
\setlength{\tabcolsep}{6pt}
\renewcommand{\arraystretch}{1.15}

\begin{tabularx}{\textwidth}{l Y Y r r l}
\toprule
Specialty & Region / Procedure & Sources & \#Videos & Hours & Public/Private \\
\midrule
Neurosurgery &
Endonasal pituitary; cranial; spine &
PitVis~\cite{das2025pitvis}; Web+\allowbreak Institutional &
2,303 & 2,860.85 & Mixed \\
\addlinespace

General surgery &
Abdominal (various) &
AVOS~\cite{goodman2024avos} &
10,874 & 125.16 & Public \\
\addlinespace

Ophthalmology &
Cataract / ophthalmic &
OphNet~\cite{hu2024ophnet}; CATARACTS~\cite{alhajj2019cataracts} &
1,969 & 230.76 & Public \\
\addlinespace

Gastric surgery &
Gastric bypass; ESD &
MultiBypass140~\cite{lavanchy2024multibypass}; CoPESD~\cite{wang2024copesd} &
180 & 252.10 & Public \\
\addlinespace

Laparoscopic cholecystectomy &
Cholecystectomy &
Cholec80~\cite{twinanda2016endonet}; M2CAI-2016~\cite{stauder2017tumlap} &
121 & 77.51 & Public \\
\addlinespace

Colonoscopy &
Endoscopy &
EndoFM~\cite{wang2023endofm} &
30,000 & 45.00 & Public \\
\addlinespace

Gynecology &
Hysterectomy &
AutoLaparo~\cite{wang2022autolaparo} &
25 & 23.12 & Public \\
\addlinespace

Urology &
Robot-assisted prostatectomy &
PSI-AVA~\cite{valderrama2023psiava} &
8 & 20.45 & Public \\
\addlinespace

Hepatic surgery &
Laparoscopic liver resection &
PmLR50~\cite{guo2025pmlr50} &
50 & 6.95 & Public \\
\addlinespace

Skill training &
Bench-top / training &
JIGSAWS~\cite{ahmidi2017dataset}; AIxSuture~\cite{hoffmann2024aixsuture} &
206 & 5.27 & Public \\
\addlinespace

Bronchoscopy &
Endoscopy &
Institutional &
60 & 50.00 & Private \\
\addlinespace

Breast surgery &
Egocentric / OR &
EgoSurgery~\cite{fujii2024egosurgery} &
21 & 10.55 & Mixed \\
\bottomrule
\end{tabularx}

\caption{Overview of data sources and scale for SurgMotion-15M. The dataset features a comprehensive collection spanning 12 diverse medical specialties, 13 organs, covering multiple surgical procedures and anatomical regions. It features a massive scale in terms of video count and duration (e.g., over 2,800 hours for Neurosurgery), possessing a high degree of significant diversity and temporal extent.
}
\label{tab:unisurg_sources}
\end{table*}

\subsection{The SurgMotion-15M Dataset}
\label{sec:dataset}
We pretrain on \textbf{SurgMotion-15M}, a large-scale multi-specialty surgical video collection aggregated from both public datasets and private institutional sources (Table~\ref{tab:unisurg_sources}).
SurgMotion-15M covers diverse anatomical regions and procedures, and is used to learn transferable spatiotemporal representations under strong domain redundancy and long-duration recordings. 


\subsection{Overview}
\label{sec:overview}
Our approach builds on the masked latent prediction paradigm of VJEPA, with an online branch predicting representations of masked regions and a target (EMA) branch providing stable learning targets.
We introduce three components tailored to surgical videos:
(i) a \textbf{motion-guided} latent masked prediction loss that emphasizes dynamically informative regions (e.g., tools and tool--tissue interaction),
(ii) a spatiotemporal affinity self-distillation loss that preserves relational structure without being dominated by visible/background tokens,
and (iii) a \textbf{spatiotemporal feature diversity regularization} (SFDR) loss to mitigate collapse under visually homogeneous scenes.

\begin{figure*}[t]
    \centering
    \includegraphics[width=\textwidth]{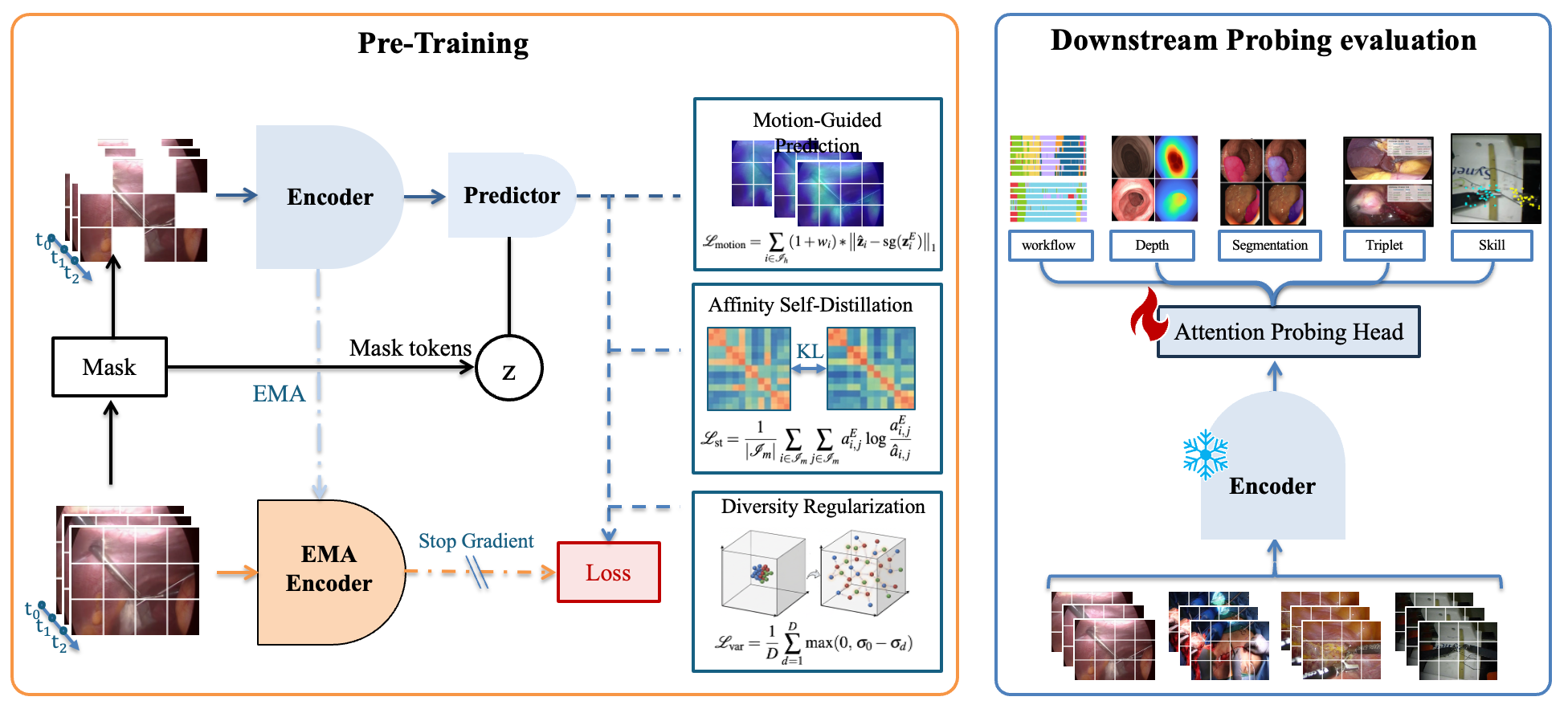}
    \caption{
    Overview of our framework.
    \textbf{Left: Pre-training.}
    A masked surgical video clip is processed by an online encoder and predictor to infer latent representations for masked tubes, while an EMA target encoder encodes the full clip to provide stable stop-gradient targets.
    Training is driven by three complementary objectives:
    motion-guided latent prediction, spatiotemporal affinity self-distillation, and spatiotemporal feature diversity regularization.
    \textbf{Right: Downstream probing evaluation.}
    After pre-training, the encoder is frozen and evaluated with lightweight probing heads on multiple surgical video understanding tasks.
    }
    \label{fig:method}
\end{figure*}

Figure~\ref{fig:method} illustrates the overall framework.
During pre-training, the online encoder processes only visible tubes, while masked tokens are fed to the predictor to infer latent representations for missing regions.
In parallel, an EMA target encoder observes the full clip and provides stop-gradient latent targets.
The three proposed objectives act on the predicted masked tokens from complementary perspectives:
The motion-guided prediction loss prioritizes dynamically salient masked tubes,
The affinity self-distillation loss preserves the relational structure among masked tokens,
and the diversity regularization prevents feature collapse in visually repetitive surgical scenes.
After pre-training, we freeze the encoder and evaluate the learned representation with lightweight probing heads on diverse downstream tasks, demonstrating its transferability across surgical video understanding settings.

\subsection{Architecture}
\label{sec:architecture}

As shown in Fig.~\ref{fig:method}, our model follows a teacher--student style masked latent prediction framework with an online branch and an EMA target branch.
The online branch receives only visible tubes and predicts latent representations for masked regions, whereas the EMA branch encodes the full clip to produce stable training targets.

\paragraph{Tube tokenization.}
Given a clip \(\mathbf{X}\in\mathbb{R}^{T\times H\times W\times C}\), each frame is partitioned into non-overlapping patches of size \(16\times16\).
A \emph{tube} corresponds to a fixed spatial patch location across all \(T\) frames, yielding
\[
N=\frac{H}{16}\cdot\frac{W}{16}
\]
tube tokens per clip.
Let \(\mathcal{I}_m\) and \(\mathcal{I}_v\) denote masked and visible tube indices.

\paragraph{Online branch.}
The context encoder \(c_{\theta}\) encodes only visible tubes to produce visible latents
\[
\mathbf{Z}_v = c_{\theta}(\mathbf{X};\mathcal{I}_v)\in\mathbb{R}^{|\mathcal{I}_v|\times D}.
\]
For each masked tube \(i\in\mathcal{I}_m\), we initialize a learnable masked token
\[
\mathbf{m}_i = \mathbf{q} + \mathbf{e}_i,
\]
where \(\mathbf{q}\in\mathbb{R}^{D}\) is a shared learnable query and \(\mathbf{e}_i\) is the positional embedding.
A masked token predictor \(p_{\phi}\) then takes the visible latents and masked tokens as input and outputs predicted latents for masked tubes:
\[
[\tilde{\mathbf{Z}}_v;\hat{\mathbf{Z}}_m] = p_{\phi}([\mathbf{Z}_v;\mathbf{M}_m]),
\quad
\hat{\mathbf{Z}}_m\in\mathbb{R}^{|\mathcal{I}_m|\times D}.
\]

\paragraph{Target branch (EMA).}
An EMA encoder \(c_{\bar{\theta}}\) (with the same architecture as \(c_{\theta}\)) encodes the full clip to produce target latents
\[
\mathbf{Z}^{E} = c_{\bar{\theta}}(\mathbf{X})\in\mathbb{R}^{N\times D}.
\]
All targets use stop-gradient, denoted \(\mathrm{sg}(\cdot)\).
This branch provides the stable targets shown in Fig.~\ref{fig:method}, preventing representation drift during pre-training.

\subsection{Training Objectives}
\label{sec:objectives}

As depicted in Fig.~\ref{fig:method}, the three losses supervise the predicted masked tokens from complementary aspects: \(\mathcal{L}_{\mathrm{motion}}\) focuses learning on dynamic surgical content, \(\mathcal{L}_{\mathrm{st}}\) preserves pairwise spatiotemporal relations among masked tokens, and \(\mathcal{L}_{\mathrm{var}}\) encourages a non-collapsed feature space.

\subsubsection{Motion-Guided Latent Masked Prediction}
\label{sec:motion_loss}
Surgical videos are temporally redundant; clinically meaningful changes are often concentrated around tools and interaction regions. We therefore upweight masked tubes that exhibit stronger spatiotemporal motion change in pixel space, corresponding to the \emph{Motion-Guided Prediction} block in Fig.~\ref{fig:method}.

\paragraph{Motion gradient.}
Let \(\mathbf{x}_{t,i}\) be the vectorized pixel patch of tube \(i\) at time \(t\).
Let \(\mathcal{N}(i)\) be the 4-neighborhood of tube \(i\) on the spatial patch grid.
We compute a motion score \(g_i\) as
\[
g_i
=
\frac{1}{T-1}\sum_{t=1}^{T-1}\left\|\mathbf{x}_{t+1,i}-\mathbf{x}_{t,i}\right\|_{1}
+
\frac{1}{T}\sum_{t=1}^{T}\frac{1}{|\mathcal{N}(i)|}\sum_{j\in\mathcal{N}(i)}\left\|\mathbf{x}_{t,i}-\mathbf{x}_{t,j}\right\|_{1}.
\]

\paragraph{Top-\(K\) emphasis among masked tubes.}
Among masked tubes, we select the top \(K=3\) by motion gradient:
\[
\mathcal{I}_h = \mathrm{TopK}\big(\{g_i\}_{i\in\mathcal{I}_m}, K\big).
\]
We assign soft weights with \(\gamma=2\):
\[
w_i = \frac{\exp(\gamma g_i)}{\sum_{k\in\mathcal{I}_h}\exp(\gamma g_k)},
\qquad i\in\mathcal{I}_h.
\]

\paragraph{Motion-guided prediction loss.}
For masked tube \(i\), let \(\hat{\mathbf{z}}_i\) be the predicted latent and \(\mathbf{z}^E_i\) the EMA target latent.
We use a weighted \(\ell_1\) regression:
\[
\mathcal{L}_{\mathrm{motion}}
=
\sum_{i\in\mathcal{I}_h}
(1+w_i)
\left\|
\hat{\mathbf{z}}_{i} - \mathrm{sg}(\mathbf{z}^{E}_{i})
\right\|_{1}.
\]
This design biases learning toward masked regions with stronger motion cues, which in surgical videos often correspond to active tools or tool--tissue interactions.

\subsubsection{Spatiotemporal Affinity Self-Distillation}
\label{sec:st_distill}
To preserve relational structure while avoiding dominance by visible/background tokens, we distill tube-to-tube affinity distributions \emph{within masked tubes only}, as illustrated by the \emph{Affinity Self-Distillation} block in Fig.~\ref{fig:method}.
Let \(\mathrm{norm}(\mathbf{u})=\mathbf{u}/\max(\|\mathbf{u}\|_2,\epsilon)\).
Define normalized masked latents
\[
\hat{\mathbf{u}}_i = \mathrm{norm}(\hat{\mathbf{z}}_i),\qquad
\mathbf{u}^E_i = \mathrm{norm}(\mathbf{z}^E_i).
\]
For each masked tube \(i\in\mathcal{I}_m\), we compute affinity distributions with temperature \(\tau\):
\[
\hat{a}_{i,j} = \frac{\exp(\hat{\mathbf{u}}_i^{\top}\hat{\mathbf{u}}_j/\tau)}
{\sum_{k\in\mathcal{I}_m}\exp(\hat{\mathbf{u}}_i^{\top}\hat{\mathbf{u}}_k/\tau)},
\quad
a^{E}_{i,j} = \frac{\exp((\mathbf{u}^E_i)^{\top}(\mathbf{u}^E_j)/\tau)}
{\sum_{k\in\mathcal{I}_m}\exp((\mathbf{u}^E_i)^{\top}(\mathbf{u}^E_k)/\tau)}.
\]
We distill using KL divergence:
\[
\mathcal{L}_{\mathrm{st}}
=
\frac{1}{|\mathcal{I}_m|}
\sum_{i\in\mathcal{I}_m}
\sum_{j\in\mathcal{I}_m}
a^{E}_{i,j}\,
\log\frac{a^{E}_{i,j}}{\hat{a}_{i,j}}.
\]
By matching affinity structure rather than only individual tokens, the model learns more coherent spatiotemporal organization among masked regions.

\subsubsection{Spatiotemporal Feature Diversity Regularization (SFDR)}
\label{sec:var_reg}
To reduce collapse under visually homogeneous segments, we regularize the variance of predicted masked latents, corresponding to the \emph{Diversity Regularization} block in Fig.~\ref{fig:method}.
Let \(\hat{\mathbf{Z}}_m\) be predicted masked latents aggregated over the batch, and let \(\sigma_d\) be the standard deviation of dimension \(d\).
We use
\[
\mathcal{L}_{\mathrm{var}}
=
\frac{1}{D}
\sum_{d=1}^{D}
\max(0,\, \sigma_0 - \sigma_d).
\]
This term encourages sufficient dispersion across feature dimensions and stabilizes representation learning in repetitive surgical scenes.

\subsubsection{Overall objective}
\label{sec:overall_obj}
Combining the three terms, the final training objective is
\[
\mathcal{L}
=
\mathcal{L}_{\mathrm{motion}}
+
\lambda_{\mathrm{st}}\mathcal{L}_{\mathrm{st}}
+
\lambda_{\mathrm{var}}\mathcal{L}_{\mathrm{var}},
\]
with \(\lambda_{\mathrm{st}}=0.1\) and \(\lambda_{\mathrm{var}}=0.3\).
As summarized in Fig.~\ref{fig:method}, these objectives jointly improve masked latent prediction by emphasizing dynamic content, preserving masked-token relations, and maintaining feature diversity.

\section{Experiments}
\label{sec:exp}

In this section, we rigorously evaluate SurgMotion against major state-of-the-art (SOTA) foundation models. 

\subsection{Experimental Setup}

\paragraph{Comparison Models.} We benchmark SurgMotion against 13 foundation models spanning general vision and surgical-specific domains. For general vision, we include DINOv3~\cite{seitzer2025dinov3} and VideoMAE~\cite{tong2022videomae, wang2023videomae} at multiple scales (Large and Giant/Huge). For surgical foundation models, we select the most recent and performant methods: image-based approaches EndoViT~\cite{batic2024endovit} and GastroNet~\cite{jong2025gastronet}; video-based methods EndoFM~\cite{wang2023endofm}, EndoSSL~\cite{hirsch2023self}, EndoMamba~\cite{tian2025endomamba}, and GSViT~\cite{schmidgall2024general}; scaled self-supervised models SurgeNetXL~\cite{jaspers2025scaling} and SelfSupSurg~\cite{ramesh2023dissecting}; and vision-language model SurgVLP~\cite{yuan2025surgvlp}. This selection covers diverse pretraining paradigms, including masked reconstruction, contrastive learning, predictive architectures, and multimodal alignment.

\paragraph{Evaluation Tasks.} We evaluate SurgMotion across the full spectrum of surgical video understanding. For \textbf{temporal understanding}, we assess surgical workflow recognition on 8 datasets spanning laparoscopic, neurosurgical, ophthalmic, and open surgery domains, reporting Accuracy, F1-score, and Jaccard index. For \textbf{fine-grained action understanding}, we evaluate action triplet recognition on CholecT50~\cite{nwoye2022rendezvous} (mean Average Precision for instruments, verbs, targets, and their combinations), action recognition on SurgicalActions160~\cite{schoeffmann2018surgicalactions}, AVOS~\cite{goodman2024avos}, and PolypDiag~\cite{tian2022contrastive} (Accuracy, F1), and skill assessment on JIGSAWS~\cite{ahmidi2017dataset} (MAE, Spearman correlation). For \textbf{dense visual perception}, we benchmark polyp segmentation on five colonoscopy datasets~\cite{jha2020kvasir,bernal2015cvc,vazquez2017cvc300,bernal2012cvccolon,silva2014etis} (Dice, MAE) and depth estimation on C3VD~\cite{bobrow2023c3vd} (RMSE, AbsRel, SqRel, $\delta < 1.1$).

\paragraph{Pretraining Details.}
For pretraining, each video frame is resized to have a shorter side of 256 pixels and randomly cropped to \(224\times224\) during training. All videos are processed at 1 fps. The pretraining follows a two-stage procedure: \textbf{Stage~1} uses \(T=16\) frames for 200 epochs with a learning rate of \(5\times10^{-4}\), while \textbf{Stage~2} continues from the Stage~1 checkpoint using \(T=64\) frames for 80 epochs. Balanced sampling across specialties and datasets is ensured by assigning each sample \(x\) from dataset \(d\) under specialty \(s\) a weight \(w(x)\propto 1/(N_d\,|\mathcal{D}_s|)\), equalizing sampling mass across specialties and datasets. Tube-level masking is applied with a random mask ratio \(\rho\sim\mathcal{U}(0.75, 0.95)\). Optimization is performed using AdamW with a weight decay of 0.04 and a cosine learning-rate schedule. We maintain an EMA momentum of \(\alpha=0.99925\) and use a distillation temperature of \(\tau=0.1\). Training is conducted on 4 NVIDIA H800 GPUs, with per-GPU batch sizes of 64 and 16 for Stage~1 and Stage~2, resulting in global batch sizes of 256 and 64, respectively.

\paragraph{Finetuning Details.}
To evaluate the performance of the foundation models in our study, we perform attentive probing following the procedure layed out in~\cite{assran2025vjepa2}. This involves adding an attentive probing head to the foundation backbone model, which is finetuned on the respective downstream task. The foundation model weights itself stay frozen. The attention head consists of 4 attention blocks with 16 cross-attention heads. Hyperparameter search is implemented by training 16 attentive heads with different learning parameters simultaneously and reporting the best result. We repeat this procedure for all reported foundation models.

\subsection{Surgical Workflow Recognition}

We evaluate temporal understanding through surgical workflow recognition, the task of identifying the current surgical phase from video. This capability is fundamental for applications such as real-time decision support, operating room scheduling, and automated documentation. We assess SurgMotion on 8 datasets spanning laparoscopic, open, endonasal, neurosurgical, and ophthalmic procedures.

\begin{table*}[!t]
    \centering
    \caption{\textbf{Surgical workflow recognition on laparoscopic procedures.} Evaluation on cholecystectomy (Cholec80, M2CAI-2016) and hysterectomy (AutoLaparo). We report Accuracy (Acc), F1-score (F1), and Jaccard index. Best results are marked in \textbf{bold}.}
    \label{tab:sota_laparoscopy}
    \small
    \setlength{\tabcolsep}{7pt}
    \renewcommand{\arraystretch}{1.2}
    \begin{tabular}{l|ccc|ccc|ccc}
        \toprule
        \multirow{2}{*}{\textbf{Model}} & \multicolumn{3}{c|}{\textbf{Cholec 80}} & \multicolumn{3}{c|}{\textbf{AutoLaparo}} & \multicolumn{3}{c}{\textbf{M2CAI-2016}} \\
         & Acc & F1 & Jaccard & Acc & F1 & Jaccard & Acc & F1 & Jaccard \\
        \midrule
        Dino v3-L & 87.20 & 81.19 & 69.60 & 82.00 & 74.55 & 66.10 & 85.55 & 74.56 & 65.39 \\
        Dino v3-H & 86.25 & 78.64 & 68.84 & 76.12 & 65.64 & 55.03 & 80.01 & 68.92 & 58.65 \\
        VideoMAE-L & 85.13 & 77.64 & 67.00 & 74.26 & 64.04 & 52.97 & 77.24 & 65.72 & 54.55 \\
        VideoMAE-G & 79.48 & 67.51 & 54.86 & 63.40 & 52.97 & 41.66 & 60.73 & 46.75 & 35.71 \\
        EndoViT & 56.41 & 36.78 & 26.92 & 51.26 & 42.95 & 31.76 & 47.32 & 27.68 & 19.61 \\
        GastroNet & 88.90 & 81.22 & 72.02 & 83.73 & 74.25 & 65.07 & 85.53 & 73.09 & 63.90 \\
        GSViT & 39.10 & 8.06 & 5.71 & 25.18 & 6.15 & 3.88 & 36.02 & 8.52 & 6.35 \\
        Endo-Mamba & 68.81 & 54.69 & 42.58 & 80.96 & 71.74 & 62.11 & 60.00 & 42.95 & 32.52 \\
        Endo SSL & 58.70 & 38.73 & 28.07 & 48.27 & 39.90 & 27.85 & 54.22 & 33.62 & 24.75 \\
        EndoFM & 57.25 & 36.13 & 26.42 & 56.67 & 35.77 & 47.84 & 53.47 & 34.46 & 25.32 \\
        Selfsupsurg & 48.87 & 19.98 & 14.17 & 43.89 & 26.76 & 18.72 & 45.23 & 21.03 & 15.00 \\
        SurgeNetXL & 82.37 & 69.17 & 57.95 & 68.21 & 53.05 & 43.12 & 69.87 & 55.23 & 44.04 \\
        SurgVLP & 84.46 & 74.12 & 63.09 & 74.36 & 65.57 & 54.15 & 75.85 & 61.81 & 50.49 \\
        \midrule
        \rowcolor{blue!15} \textbf{SurgMotion} & \textbf{91.05} & \textbf{84.17} & \textbf{77.95} & \textbf{86.37} & \textbf{78.73} & \textbf{69.86} & \textbf{89.45} & \textbf{82.91} & \textbf{75.42} \\
        \bottomrule
    \end{tabular}
\end{table*}

\paragraph{Laparoscopic Surgery.}
We first evaluate on established laparoscopic benchmarks: Cholec80~\cite{twinanda2016endonet} and M2CAI-2016~\cite{stauder2017tumlap} for cholecystectomy, and AutoLaparo~\cite{wang2022autolaparo} for hysterectomy. As shown in Table~\ref{tab:sota_laparoscopy}, SurgMotion consistently outperforms all baselines across all three datasets. On Cholec80, SurgMotion achieves 91.05\% accuracy and 77.95\% Jaccard, surpassing DINOv3-L by +8.35\% in Jaccard index. This substantial improvement in temporal overlap indicates that SurgMotion captures phase boundaries more precisely than frame-wise discriminative models. On AutoLaparo, where several baselines struggle with complex multi-stage workflows (e.g., GSViT achieves only 25.18\% accuracy), SurgMotion maintains robust performance at 86.37\% accuracy. On M2CAI-2016, SurgMotion achieves 89.45\% accuracy and 75.42\% Jaccard, outperforming the next best method by over 3\% in F1-score.

\begin{table*}[!t]
    \centering
    \caption{\textbf{Surgical workflow recognition on open, endonasal, and neurosurgical procedures.} Evaluation on egocentric open surgery (EgoSurgery), endonasal pituitary surgery (PitVis), and cranial neurosurgery (Atlas-Neurosurgical). Best results are marked in \textbf{bold}.} 
    \label{tab:sota_gen}
    \small
    \setlength{\tabcolsep}{10pt}
    \setlength{\tabcolsep}{3pt}
    \renewcommand{\arraystretch}{1.2}
    \begin{tabular}{l|ccc|ccc|ccc}
        \toprule
        \multirow{2}{*}{\textbf{Model}} & \multicolumn{3}{c|}{\textbf{EgoSurgery}} & \multicolumn{3}{c|}{\textbf{PitVis}} & \multicolumn{3}{c}{\textbf{Atlas-Neurosurgical}} \\
         & Acc & F1-score & Jaccard & Acc & F1-score & Jaccard & Acc & F1-score & Jaccard \\
        \midrule
        DinoV3-L & 75.37 & 32.43 & 26.03 & 78.24 & 55.75 & 46.24 & 79.31 & 63.19 & 56.96 \\
        DinoV3-H & 70.08 & 36.34 & 29.03 & 76.70 & 50.18 & 41.09 & 76.54 & 59.76 & 53.70 \\
        VideoMAE-L & 64.52 & 26.84 & 19.82 & 80.62 & 56.89 & 47.75 & 76.55 & 59.13 & 52.13 \\
        VideoMAE-G & 62.25 & 20.40 & 16.13 & 72.52 & 44.34 & 35.59 & 77.06 & 53.89 & 47.08 \\
        EndoViT & 32.32 & 9.29 & 6.31 & 53.09 & 28.32 & 20.77 & 73.71 & 48.51 & 42.36 \\
        GastroNet & 48.12 & 16.08 & 12.87 & 82.95 & 65.22 & 55.85 & 77.79 & 58.71 & 51.94 \\
        GSViT & 35.74 & 7.33 & 5.17 & 26.68 & 3.42 & 2.29 & 61.36 & 28.01 & 24.15 \\
        Endo Mamba & 53.30 & 13.90 & 10.32 & 61.54 & 35.27 & 27.84 & 70.37 & 44.57 & 37.87 \\
        EndoSSL & 49.78 & 12.04 & 8.86 & 56.72 & 27.00 & 19.60 & 72.14 & 48.20 & 41.86 \\
        EndoFM & 54.85 & 36.13 & 26.42 & 63.31 & 31.86 & 23.46 & 77.87 & 46.59 & 41.09 \\
        Selfsupsurg & 36.03 & 7.38 & 5.22 & 47.87 & 18.63 & 13.43 & 65.19 & 37.26 & 32.18 \\
        SurgeNetXL & 35.82 & 7.34 & 5.18 & 50.99 & 18.60 & 13.83 & 69.36 & 44.67 & 38.35 \\
        SurgVLP & 57.44 & 23.26 & 17.46 & 70.99 & 29.89 & 32.06 & 72.56 & 51.16 & 44.52 \\
        \midrule
        \rowcolor{blue!15} \textbf{SurgMotion} & \textbf{75.57} & \textbf{50.72} & \textbf{42.95} & \textbf{86.52} & \textbf{75.53} & \textbf{65.51} & \textbf{83.48} & \textbf{65.26} & \textbf{59.35} \\
        \bottomrule
    \end{tabular}
\end{table*}

\paragraph{Open, Endonasal, and Neurosurgery.}
We evaluate on three distinct surgical settings beyond laparoscopy: EgoSurgery~\cite{fujii2024egosurgery} for egocentric open surgery, PitVis~\cite{das2025pitvis} for endonasal pituitary surgery, and Atlas-Neurosurgical for cranial neurosurgery. As shown in Table~\ref{tab:sota_gen}, SurgMotion demonstrates substantial improvements over all baselines. On EgoSurgery, which features egocentric viewpoints and variable lighting conditions, SurgMotion achieves 50.72\% F1-score, outperforming DINOv3-L by +18.3\% and the best surgical foundation model (EndoFM) by +14.6\%. On PitVis, SurgMotion attains 75.53\% F1-score, surpassing GastroNet by +10.3\%. On Atlas-Neurosurgical, SurgMotion achieves 65.26\% F1-score, outperforming all baselines including DINOv3-L (+2.1\%) and the surgical-specific SurgVLP (+14.1\%).

\begin{table*}[!t]
    \centering
    \caption{\textbf{Surgical workflow recognition on ophthalmic and hepatic procedures.} Evaluation on ophthalmic surgery (OphNet) and laparoscopic liver resection (PMLR50). Best results are marked in \textbf{bold}.}
    \label{tab:sota_specialized}
    \small
    \setlength{\tabcolsep}{13pt}
    \renewcommand{\arraystretch}{1.2}
    \begin{tabular}{l|ccc|ccc}
        \toprule
        \multirow{2}{*}{\textbf{Model}} & \multicolumn{3}{c|}{\textbf{OphNet}} & \multicolumn{3}{c}{\textbf{PMLR 50}} \\
         & Acc & F1-score & Jaccard & Acc & F1-score & Jaccard \\
        \midrule
        DinoV3-L & 62.80 & 39.99 & 31.99 & 82.98 & 63.85 & 53.76 \\
        DinoV3-H & 63.31 & 38.93 & 31.68 & 73.18 & 37.59 & 29.74 \\
        VideoMAE-L & 58.97 & 34.35 & 27.00 & 84.50 & 68.68 & 58.25 \\
        VideoMAE-G & 51.53 & 27.98 & 21.38 & 84.07 & 65.61 & 54.94 \\
        EndoViT & 28.14 & 9.57 & 6.91 & 75.41 & 41.92 & 34.51 \\
        GastroNet & 61.41 & 37.95 & 30.56 & 78.28 & 56.94 & 47.23 \\
        GSViT & 17.81 & 2.52 & 1.69 & 65.37 & 16.13 & 14.40 \\
        Endo Mamba & 28.94 & 10.37 & 7.53 & 75.50 & 46.34 & 35.91 \\
        EndoSSL & 28.33 & 9.73 & 6.86 & 78.16 & 46.43 & 39.88 \\
        EndoFM & 27.88 & 8.06 & 5.66 & 65.44 & 35.35 & 25.03 \\
        Selfsupsurg & 23.15 & 6.36 & 4.48 & 70.55 & 29.81 & 24.73 \\
        SurgeNetXL & 24.97 & 7.69 & 5.79 & 65.26 & 16.13 & 13.39 \\
        SurgVLP & 48.13 & 24.61 & 18.76 & 76.75 & 51.23 & 42.38 \\
        \midrule
        \rowcolor{blue!15} \textbf{SurgMotion} & \textbf{73.04} & \textbf{51.22} & \textbf{43.83} & \textbf{91.91} & \textbf{85.65} & \textbf{78.56} \\
        \bottomrule
    \end{tabular}
\end{table*}

\paragraph{Ophthalmic and Hepatic Surgery.}
We further evaluate on OphNet~\cite{hu2024ophnet} for ophthalmic procedures and PMLR50~\cite{guo2025pmlr50} for laparoscopic liver resection. As shown in Table~\ref{tab:sota_specialized}, SurgMotion achieves strong performance in both domains. On OphNet, which features a very diverse set of opthalmic procedures with over 33 distinc phases, SurgMotion attains 73.04\% accuracy and 51.22\% F1-score, exceeding DINOv3-L by over 10\% in both metrics. On PMLR50, SurgMotion achieves 91.91\% accuracy and 85.65\% F1-score, outperforming VideoMAE-L by +7.4\% in accuracy and +17.0\% in F1-score.

\subsection{Fine-Grained Action Understanding}

Beyond phase-level workflow recognition, surgical video understanding requires fine-grained reasoning about actions, instruments, and their interactions. We evaluate SurgMotion on three complementary tasks: action recognition, action triplet recognition, and skill assessment.

\begin{table*}[!t]
    \centering
    \caption{\textbf{Action recognition.} Evaluation on laparoscopic actions (SurgicalActions160), open surgery (AVOS), and colonoscopic diagnosis (PolypDiag). We report Accuracy (Acc) and F1-score (F1). Best results are marked in \textbf{bold}.}
    \label{tab:sota_action}
    \small
    \setlength{\tabcolsep}{13pt}
    \renewcommand{\arraystretch}{1.2}
    \begin{tabular}{l|cc|cc|cc}
        \toprule
        \multirow{2}{*}{\textbf{Model}} & \multicolumn{2}{c|}{\textbf{SurgicalActions160}} & \multicolumn{2}{c|}{\textbf{AVOS}} & \multicolumn{2}{c}{\textbf{PolypDiag}} \\
         & Acc & F1-score & Acc & F1-score & Acc & F1-score \\
        \midrule
        Dino V3-L & 64.58 & 63.78 & 64.27 & 26.02 & 95.00 & 93.54 \\
        Dino V3-H & 75.00 & 72.35 & 77.37 & 45.12 & 97.50 & 96.67 \\
        VideoMAE-L & 70.83 & 70.55 & 59.60 & 23.01 & 96.25 & 94.91 \\
        VideoMAE-G & 41.67 & 40.67 & 48.34 & 8.15 & 97.50 & 96.55 \\
        EndoViT & 16.67 & 10.25 & 48.34 & 8.15 & 96.25 & 94.91 \\
        GastroNet & 41.67 & 41.61 & 54.54 & 17.84 & 98.75 & 98.30 \\
        GSViT & 18.75 & 11.06 & 48.34 & 8.15 & 75.00 & 42.86 \\
        Endo Mamba & 58.45 & 36.23 & 31.55 & 22.86 & 23.54& 20.05\\
        EndoSSL & 12.50 & 6.77 & 48.34 & 8.15 & 92.50 & 90.31 \\
        EndoFM & 8.33 & 3.80 & 48.34 & 8.15 & 97.50 & 96.44 \\
        Selfsupsurg & 10.42 & 7.93 & 48.34 & 8.15 & 90.00 & 85.06 \\
        SurgNetXL & 50.00 & 49.08 & 50.44 & 13.50 & 97.50 & 96.67 \\
        SurgVLP & 39.58 & 36.64 & 49.66 & 11.63 & 93.75 & 91.80 \\
        \midrule
        \rowcolor{blue!15} \textbf{SurgMotion} & \textbf{75.63} & \textbf{72.62} & \textbf{70.52} & \textbf{66.32} & \textbf{98.81} & \textbf{98.34} \\
        \bottomrule
    \end{tabular}
\end{table*}

\paragraph{Action Recognition.}
We evaluate short-term action recognition on SurgicalActions160~\cite{schoeffmann2018surgicalactions} for laparoscopic actions, AVOS~\cite{goodman2024avos} for open surgery, and PolypDiag~\cite{tian2022contrastive} for colonoscopic polyp diagnosis. As shown in Table~\ref{tab:sota_action}, SurgMotion achieves competitive or superior performance across all three benchmarks. On SurgAction-160, SurgMotion achieves 75.63\% accuracy, matching DINOv3-H while substantially outperforming surgical-specific models such as SurgeNetXL (50.00\%) and SurgVLP (39.58\%). On AVOS, SurgMotion attains 66.32\% F1-score, dramatically improving over all baselines including DINOv3-H (45.12\% F1) by +21.2\%. On PolypDiag, SurgMotion achieves 98.81\% accuracy, slightly exceeding the strong GastroNet baseline (98.75\%).

\begin{table*}[!t]
    \centering
    \caption{\textbf{Action triplet recognition on CholecT50.} We report Average Precision (AP) for individual components—Instruments (I), Verbs (V), Targets (T)—and their associations (IV, IT, IVT). Best results are in \textbf{bold}.}
    \label{tab:triplet_full}
    \renewcommand{\arraystretch}{1.2}
    \small
    \setlength{\tabcolsep}{13pt}
    \begin{tabular}{l|ccc|ccc}
        \toprule
        \multirow{2}{*}{\textbf{Model}} & \multicolumn{3}{c|}{\textbf{Component detection}} & \multicolumn{3}{c}{\textbf{Triplet association}} \\
         & AP-I & AP-V & AP-T & AP-IV & AP-IT & AP-IVT \\
        \midrule
        Dino v3-L & 87.16 & 53.90 & 43.44 & 37.31 & 40.81 & 36.85 \\
        Dino v3-H & 90.18 & 55.83 & 42.70 & 36.96 & 34.09 & 30.06 \\
        VideoMAE-G & 77.06 & 46.69 & 40.42 & 30.60 & 29.08 & 24.41 \\
        VideoMAE & 81.58 & 47.93 & 36.78 & 31.11 & 30.40 & 25.72 \\
        EndoViT & 52.41 & 32.01 & 27.45 & 20.65 & 19.42 & 18.05 \\
        GastroNet & 73.69 & 48.14 & 39.44 & 31.22 & 30.05 & 27.37 \\
        GSViT & 18.65 & 12.17 & 8.40 & 6.36 & 3.56 & 3.20 \\
        Endo Mamba & 58.45 & 36.23 & 31.55 & 22.86 & 23.54 & 20.05 \\
        EndoSSL & 50.47 & 34.79 & 27.93 & 22.67 & 20.20 & 20.17 \\
        EndoFM & 48.57 & 28.52 & 27.51 & 18.44 & 19.71 & 18.98 \\
        Selfsupsurg & 45.50 & 29.16 & 22.65 & 18.25 & 14.41 & 13.50 \\
        SurgNetXL & 74.64 & 45.58 & 40.08 & 29.92 & 32.51 & 27.36 \\
        SurgVLP & 75.56 & 44.75 & 33.29 & 29.30 & 28.26 & 24.11 \\
        \midrule
        \rowcolor{blue!15} \textbf{SurgMotion} & \textbf{91.55} & \textbf{57.72} & \textbf{48.18} & \textbf{40.39} & \textbf{43.47} & \textbf{39.54} \\
        \bottomrule
    \end{tabular}
\end{table*}

\paragraph{Action Triplet Recognition.}
We evaluate the model's ability to disentangle complex surgical interactions on CholecT50~\cite{nwoye2022rendezvous}, which requires simultaneous recognition of Instruments (I), Verbs (V), and Targets (T). As shown in Table~\ref{tab:triplet_full}, SurgMotion achieves state-of-the-art performance across all metrics. For individual components, SurgMotion attains 91.55\% AP-I, 57.72\% AP-V, and 48.18\% AP-T, outperforming DINOv3-L by +4.4\% on instruments, +3.8\% on verbs, and +4.7\% on targets. For the challenging triplet association task (AP-IVT), SurgMotion achieves 39.54\%, surpassing DINOv3-L by +2.7\%. The consistent improvement in verb recognition confirms that SurgMotion's video-native pretraining captures motion dynamics critical for distinguishing fine-grained surgical actions.

\begin{table}[!t]
    \centering
    \caption{\textbf{Skill assessment on JIGSAWS.} We report Mean Absolute Error (MAE, lower is better) and Spearman correlation (higher is better). Best results are marked in \textbf{bold}.}
    \label{tab:skill}
    \small
    \setlength{\tabcolsep}{40pt}
    \renewcommand{\arraystretch}{1.2}
    \begin{tabular}{l|cc}
        \toprule
        \textbf{Model} & \textbf{MAE} & \textbf{Spearman} \\
        \midrule
        DinoV3-L & 3.9895 & 0.4159 \\
        DinoV3-H & 3.7070 & 0.4167 \\
        VideoMAE-L & 3.9189 & 0.4174 \\
        VideoMAE-G & 3.9720 & 0.3362 \\
        EndoViT & 4.1000 & 0.4431 \\
        GastroNet & 3.5962 & 0.5002 \\
        GSViT & 4.7528 & NaN \\
        EndoSSL & 4.2182 & 0.3986 \\
        EndoFM & 4.0895 & 0.4465 \\
        Selfsupsurg & 4.0229 & NaN \\
        SurgNetXL & 4.7663 & 0.0509 \\
        SurgVLP & 3.9817 & 0.4071 \\
        \midrule
        \rowcolor{blue!15} \textbf{SurgMotion} & \textbf{2.649} & \textbf{0.770} \\
        \bottomrule
    \end{tabular}
\end{table}

\paragraph{Skill Assessment.}
We evaluate surgical skill quantification on JIGSAWS~\cite{ahmidi2017dataset}, a benchmark of robotic surgery tasks with expert skill annotations. The annotations show a skill rating from 1 to 5 in 5 categories, we predict the global skill score. As shown in Table~\ref{tab:skill}, SurgMotion achieves the MAE of 2.649 and the Spearman correlation of 0.770, substantially outperforming all baselines. Compared to GastroNet (MAE 3.596, Spearman 0.500), SurgMotion reduces prediction error by 26\% while improving rank correlation by 27 percentage points. This strong performance suggests that SurgMotion's learned representations capture not only what actions are performed but also the quality and fluency of their execution.

\subsection{Dense Visual Perception}

Beyond high-level temporal and action understanding, surgical AI systems require dense pixel-level predictions for tasks such as tissue segmentation and 3D reconstruction. We evaluate SurgMotion on polyp segmentation and depth estimation to assess whether our video-level pretraining transfers to dense prediction tasks.

\begin{table}[!t]
    \centering
    \caption{\textbf{Polyp segmentation.} Evaluation on in-distribution (Kvasir, CVC-ClinicDB) and out-of-distribution (CVC-300, CVC-ColonDB, ETIS-LaribPolypDB) benchmarks. We report Dice coefficient (higher is better) and MAE (lower is better). Best results are marked in \textbf{bold}, second-best are marked by \underline{underlined}.}
    \label{tab:seg}
    \small
    \setlength{\tabcolsep}{4pt} 
    \renewcommand{\arraystretch}{1.2}
    
    \begin{tabular}{l|cc|cc|cc|cc|cc}
        \toprule
        & \multicolumn{4}{c|}{\textbf{In-Domain}} & \multicolumn{6}{c}{\textbf{Out-of-Distribution}} \\
        \cmidrule(lr){2-5} \cmidrule(lr){6-11}
        & \multicolumn{2}{c|}{\textbf{ClinicDB}} & \multicolumn{2}{c|}{\textbf{Kvasir}} & \multicolumn{2}{c|}{\textbf{CVC-300}} & \multicolumn{2}{c|}{\textbf{ColonDB}} & \multicolumn{2}{c}{\textbf{ETIS}} \\
        \textbf{Model} & \textbf{Dice} $\uparrow$ & \textbf{MAE} $\downarrow$ & \textbf{Dice} $\uparrow$ & \textbf{MAE} $\downarrow$ & \textbf{Dice} $\uparrow$ & \textbf{MAE} $\downarrow$ & \textbf{Dice} $\uparrow$ & \textbf{MAE} $\downarrow$ & \textbf{Dice} $\uparrow$ & \textbf{MAE} $\downarrow$ \\
        \midrule
        EndoFM & 0.8353 & 0.0222 & 0.8421 & 0.0438 & 0.8352 & 0.0126 & 0.6338 & 0.0485 & 0.4952 & 0.0527 \\
        EndoMamba & 0.7182 & 0.0398 & 0.6831 & 0.1017 & 0.6075 & 0.0451 & 0.4684 & 0.0809 & 0.2661 & 0.1175 \\
        EndoViT & 0.8271 & 0.0236 & 0.8533 & 0.0442 & 0.8303 & 0.0137 & 0.6383 & 0.0515 & 0.5345 & 0.0390 \\
        EndoSSL & 0.8349 & 0.0244 & 0.8823 & 0.0379 & 0.8622 & 0.0088 & 0.7270 & 0.0388 & 0.6659 & 0.0240 \\
        GSViT & 0.6218 & 0.0610 & 0.6238 & 0.1331 & 0.3633 & 0.1066 & 0.4027 & 0.1071 & 0.2692 & 0.1504 \\
        VideoMAE-G & 0.8323 & 0.0249 & 0.8218 & 0.0579 & 0.7883 & 0.0226 & 0.6391 & 0.0539 & 0.5242 & 0.0489 \\
        SelfSupSurg & 0.8623 & 0.0194 & 0.8452 & 0.0442 & 0.8355 & 0.0116 & 0.6619 & 0.0472 & 0.6057 & 0.0309 \\
        SurgNetXL & 0.8499 & 0.0196 & 0.8317 & 0.0569 & 0.7730 & 0.0238 & 0.6567 & 0.0550 & 0.5534 & 0.0652 \\
        SurgVLP & 0.8253 & 0.0260 & 0.8433 & 0.0530 & 0.8508 & 0.0131 & 0.6699 & 0.0493 & 0.6265 & 0.0384 \\
        GastroNet & 0.8611 & 0.0171 & 0.8897 & 0.0359 & 0.8250 & 0.0134 & 0.7182 & 0.0369 & 0.7108 & \textbf{0.0160} \\
        DINOv3-L & \underline{0.8692} & \underline{0.0164} & 0.8872 & \textbf{0.0318} & 0.8811 & \underline{0.0071} & 0.7646 & \underline{0.0342} & 0.7293 & 0.0197 \\
        DINOv3-H & 0.8676 & 0.0204 & \textbf{0.9019} & \underline{0.0336} & \underline{0.8912} & \textbf{0.0066} & \underline{0.7795} & 0.0345 & \underline{0.7692} & 0.0175 \\
        \midrule
        \rowcolor{blue!15} SurgMotion & \textbf{0.8694} & \textbf{0.0145} & \underline{0.8919} & 0.0352 & \textbf{0.9112} & \underline{0.0071} & \textbf{0.7985} & \textbf{0.0237} & \textbf{0.7798} & \underline{0.0163} \\
        \bottomrule
    \end{tabular}%
\end{table}

\paragraph{Polyp Segmentation.}
We evaluate polyp segmentation on five colonoscopy benchmarks following established protocols~\cite{fan2020pranet,kim2021uacanet,hu2025pranetv2}: Kvasir~\cite{jha2020kvasir} and CVC-ClinicDB~\cite{bernal2015cvc} as in-distribution (ID) datasets, and CVC-300~\cite{vazquez2017cvc300}, CVC-ColonDB~\cite{bernal2012cvccolon}, and ETIS-LaribPolypDB~\cite{silva2014etis} as out-of-distribution (OOD) datasets. As shown in Table~\ref{tab:seg}, SurgMotion achieves state-of-the-art performance, particularly on OOD datasets. On the ID benchmarks, SurgMotion attains Dice coefficients of 0.8694 (CVC-ClinicDB) and 0.8919 (Kvasir), matching the strongest baselines. On the more challenging OOD benchmarks, SurgMotion demonstrates superior generalization: 0.9112 Dice on CVC-300, 0.7985 on CVC-ColonDB, and 0.7798 on ETIS-LaribPolypDB, with the lowest MAE across all OOD datasets (0.0071, 0.0237, and 0.0163, respectively). These results indicate that SurgMotion's pretraining captures generalizable visual features for polyp morphology rather than dataset-specific artifacts.

\begin{table}[!t]
    \centering
    \caption{\textbf{Depth estimation on C3VD.} We report RMSE, AbsRel, SqRel (lower is better), and $\delta < 1.1$ accuracy (higher is better). Best results are marked in \textbf{bold}.}
    \label{tab:depth}
    \small
    \setlength{\tabcolsep}{15pt}
    \renewcommand{\arraystretch}{1.2}
    \begin{tabular}{l|cccc}
        \toprule
        \textbf{Model} & \textbf{RMSE} $\downarrow$ & \textbf{AbsRel} $\downarrow$ & \textbf{SqRel} $\downarrow$ & $\delta < 1.1 \uparrow$ \\
        \midrule
        VideoMAE-G & 2.25 & 0.062 & 0.17 & 0.81 \\
        EndoViT & 2.05 & 0.057 & 0.15 & 0.84 \\
        GSViT & 3.65 & 0.094 & 0.45 & 0.63 \\
        EndoFM & 2.16 & 0.059 & 0.16 & 0.82 \\
        DINOV3-L & 1.92 & 0.055 & 0.13 & 0.84 \\
        DINOv3-H & 1.94 & 0.057 & 0.13 & 0.83 \\
        SelfSupSurg & 3.12 & 0.094 & 0.36 & 0.64 \\
        SurgVLP & 2.94 & 0.085 & 0.32 & 0.68 \\
        SurgeNetXL & 2.88 & 0.078 & 0.29 & 0.72 \\
        EndoSSL & 2.00 & 0.058 & 0.15 & 0.82 \\
        GastroNet & 1.91 & 0.055 & 0.13 & 0.84 \\
        InternVideo-Next-L & 2.22 & 0.064 & 0.17 & 0.79 \\
        InternVideo2-1b & 2.00 & 0.059 & 0.14 & 0.82 \\
        \midrule
        \rowcolor{blue!15} \textbf{SurgMotion} & \textbf{1.88} & \textbf{0.051} & \textbf{0.13} & \textbf{0.86} \\
        \bottomrule
    \end{tabular}
\end{table}

\paragraph{Depth Estimation.}
We evaluate monocular depth estimation on the C3VD colonoscopy benchmark~\cite{bobrow2023c3vd}, which provides ground-truth depth maps from 3D reconstructions. As shown in Table~\ref{tab:depth}, SurgMotion achieves an RMSE of 1.88, outperforming all baselines, including DINOv3-L (1.92), GastroNet (1.91), and the video-based InternVideo2-1b (2.00). SurgMotion also achieves the best AbsRel (0.051) and $\delta < 1.1$ accuracy (0.86). This strong performance is notable given that SurgMotion was not explicitly trained for geometric prediction, suggesting that our latent prediction objective implicitly captures 3D spatial structure through motion cues.

\begin{figure*}[h]
    \centering
    \vspace{-0.3cm}
        \includegraphics[width=\textwidth]{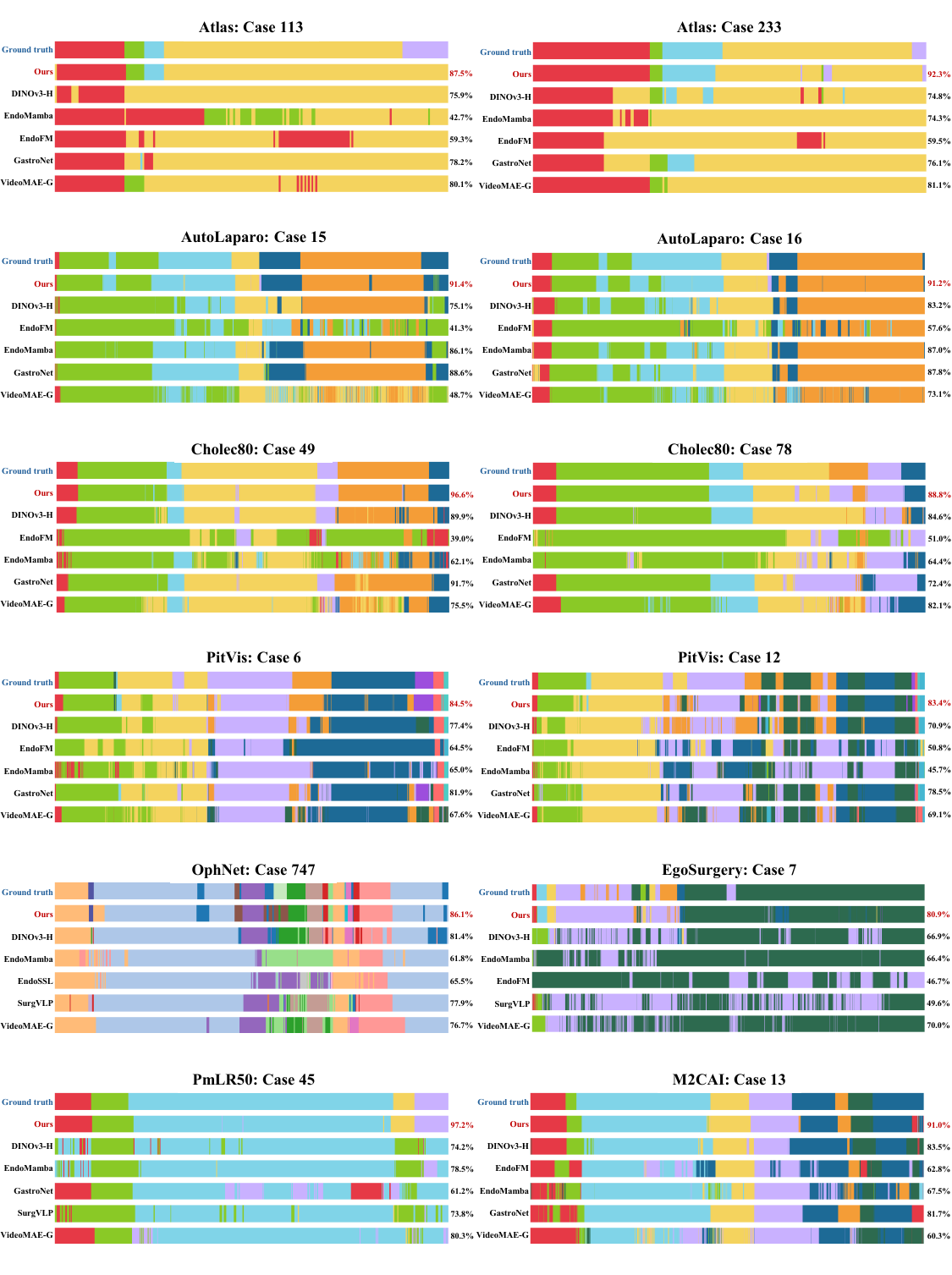} \\
    \caption{\textbf{Multi-dataset qualitative showcase of surgical phase recognition}. Frame-wise phase predictions on representative cases from different surgical datasets are compared with the ground truth and competing methods. Our method yields predictions that better match the ground truth, with clearer phase boundaries and stronger temporal consistency across diverse scenarios.}
    \label{fig:qualitative}
\end{figure*}

\begin{figure*}[h]
    \centering

    \begin{minipage}{\textwidth}
        \centering
        \includegraphics[width=\textwidth]{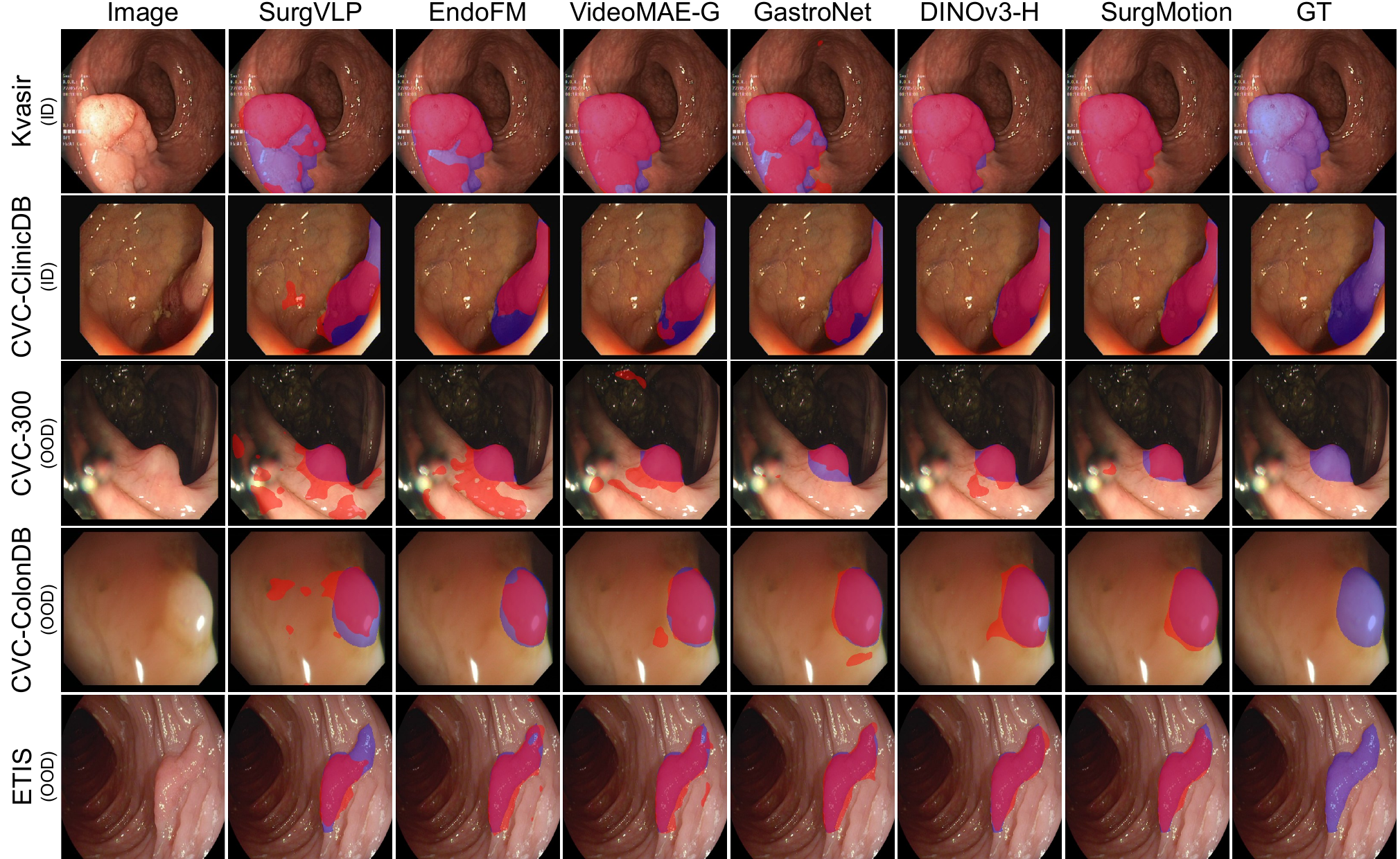}\\[-0.4em]
        \small \textbf{(a) Polyp Segmentation.} ID: Kvasir, CVC-ClinicDB. OOD: CVC-300, CVC-ColonDB, ETIS-LaribPolypDB. Predictions are overlaid in magenta; GT is overlaid in blue.
    \end{minipage}

    \vspace{0.35cm}

    \begin{minipage}{\textwidth}
        \centering
        \includegraphics[width=\textwidth]{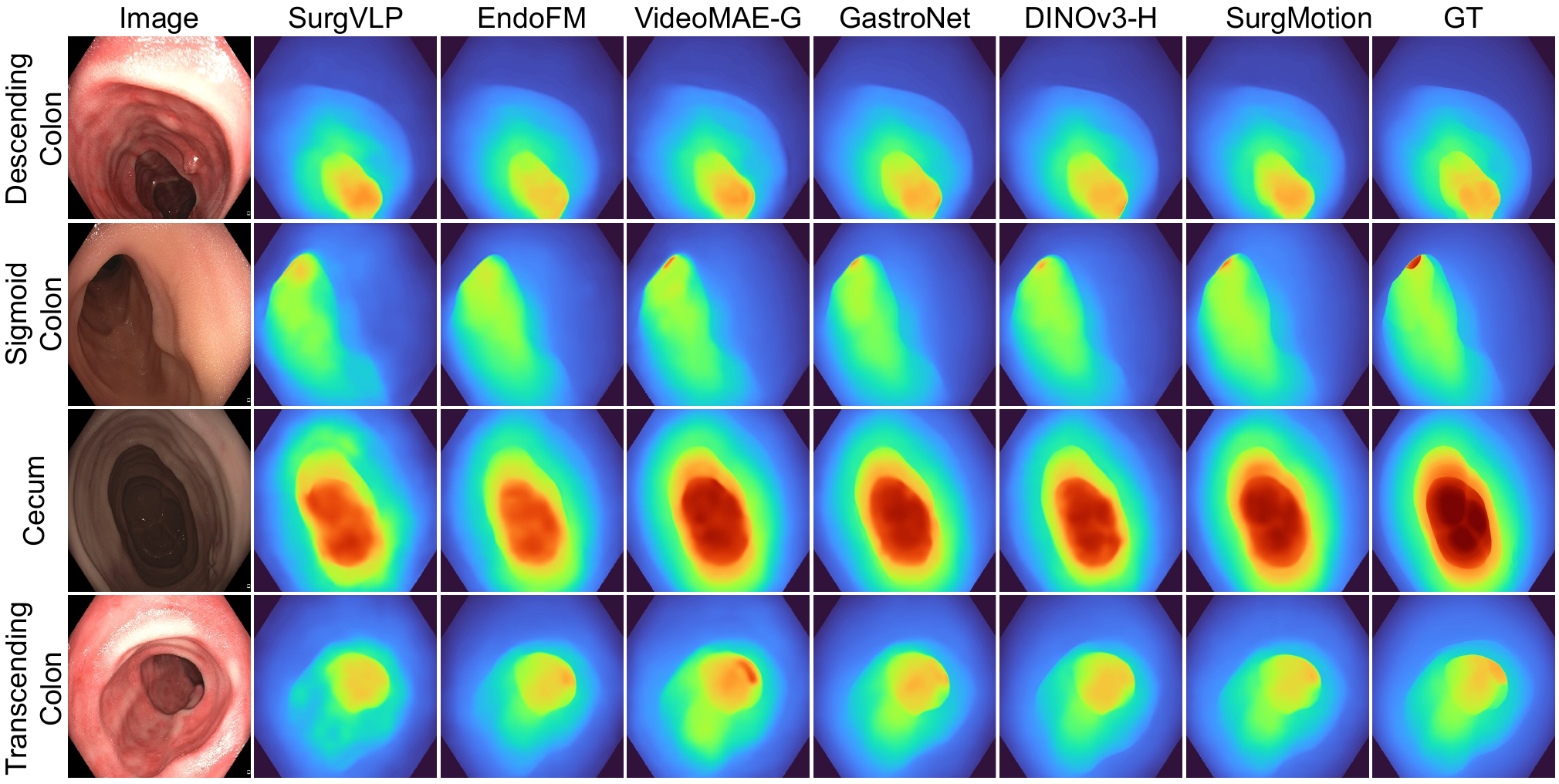}\\[-0.4em]
        \small \textbf{(b) Colonoscopic Depth Estimation (C3VD).} Depth heatmaps across colon segments (descending, sigmoid, cecum, transverse).
    \end{minipage}

    \caption{\textbf{Qualitative Results on Polyp Segmentation and Colonoscopic Depth Estimation.} Columns show the input frame, predictions from SurgVLP, EndoFM, VideoMAE-G, GastroNet, DINOv3-H, and our SurgMotion, as well as the ground truth (GT). SurgMotion yields tighter polyp boundaries under domain shift and more spatially coherent depth maps across anatomical segments.}
    \label{fig:quali_seg_depth}
\end{figure*} 

\begin{figure*}[h]
    \centering

    \begin{minipage}{\textwidth}
        \centering
        \includegraphics[width=\textwidth]{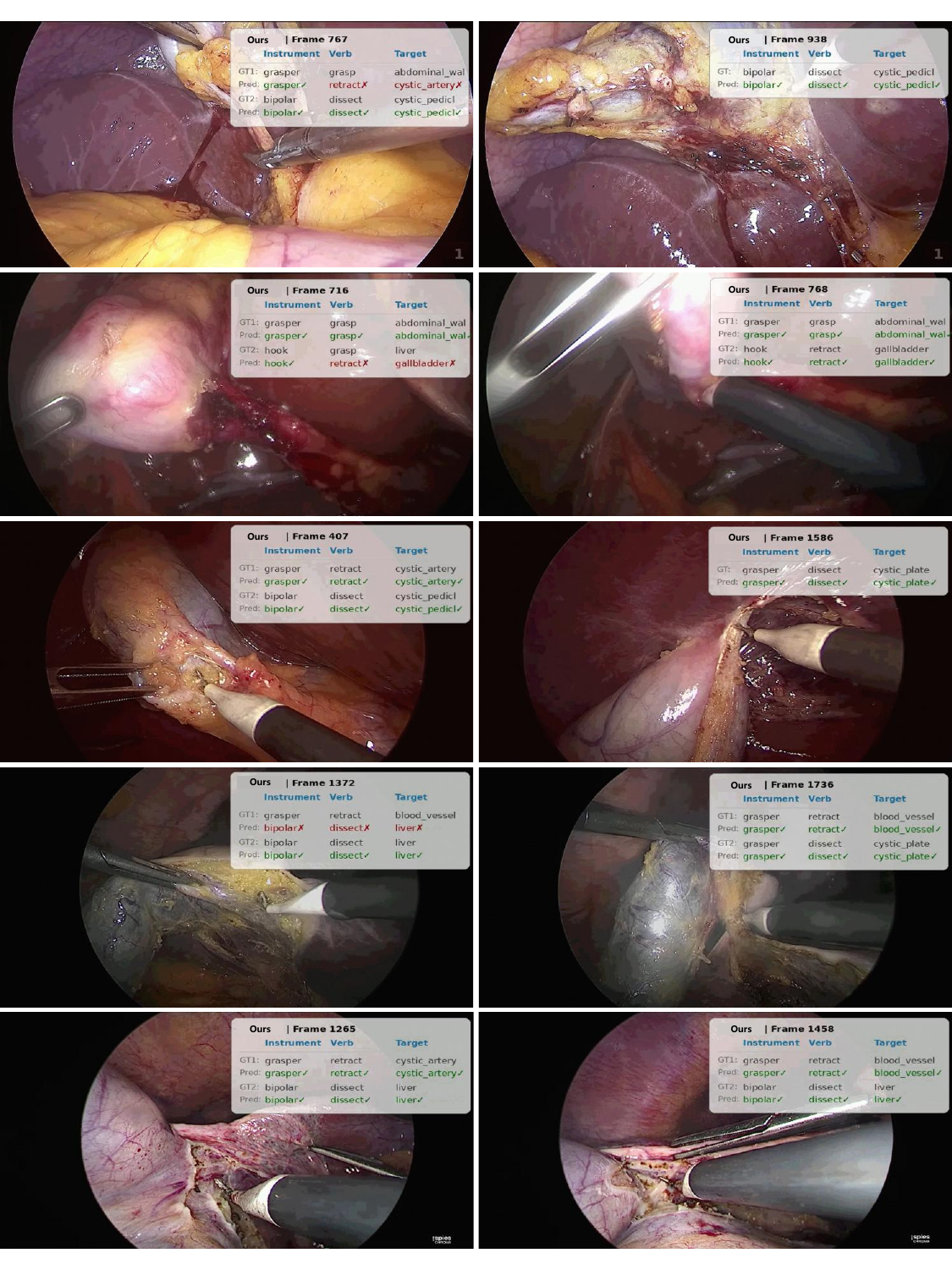}\\[-0.4em]
    \end{minipage}

    \vspace{0.35cm}

    \caption{\textbf{Illustration of surgical action triplet recognition on CholecT50 dataset.} These frames from a CholecT50 laparoscopic cholecystectomy video exemplify the model's capability to disentangle overlapping surgical actions.}
    \label{fig:quali_triplet}
\end{figure*} 

\subsection{Qualitative Analysis}

We provide qualitative visualizations to complement our quantitative evaluation and offer insight into model behavior.

\paragraph{Workflow Recognition.}
Figure~\ref{fig:qualitative} visualizes temporal phase predictions across 12 representative cases from 8 datasets. Each row shows the predicted phase sequence as a colored bar, with the ground truth at the top. SurgMotion produces smoother predictions that closely track the ground truth phase boundaries, whereas baselines exhibit frequent fragmentation and phase confusion. For instance, on Atlas Case 113, SurgMotion achieves 87.5\% accuracy while EndoMamba drops to 42.7\%, suffering from severe over-segmentation. On challenging cases such as EgoSurgery Case 7, which features dramatic viewpoint changes, SurgMotion (80.9\%) substantially outperforms EndoFM (46.7\%) and SurgVLP (49.6\%). These visualizations confirm that SurgMotion's temporal representations capture coherent phase structure rather than noisy frame-level predictions.

\paragraph{Polyp Segmentation.}
Figure~\ref{fig:quali_seg_depth}(a) shows qualitative segmentation results on both in-distribution and out-of-distribution colonoscopy datasets. SurgMotion consistently produces tighter boundaries that align closely with ground truth polyp contours. On OOD datasets (CVC-300, CVC-ColonDB, ETIS), where domain shift poses significant challenges, baselines such as VideoMAE-G and EndoFM produce incomplete or fragmented masks, while SurgMotion maintains robust segmentation. This visual evidence corroborates the quantitative OOD improvements reported in Table~\ref{tab:seg}.

\paragraph{Depth Estimation.}
Figure~\ref{fig:quali_seg_depth}(b) visualizes depth predictions across different colon segments (descending, sigmoid, cecum, transverse). SurgMotion produces spatially coherent depth maps that accurately capture the tubular geometry of the colon lumen, with smooth gradients from near (warm colors) to far (cool colors). In contrast, baselines such as SurgVLP and EndoFM exhibit noisy predictions with artifacts near tissue boundaries. The depth maps from SurgMotion most closely match the ground truth across all anatomical segments, demonstrating that our video-native pretraining captures meaningful 3D geometric structure.

\paragraph{Triplet Recognition.}
Figure~\ref{fig:quali_triplet} presents qualitative results of surgical action triplet recognition on the CholecT50 dataset. SurgMotion accurately identifies the instrument–verb–target triplets across diverse surgical contexts, such as grasping, retracting, and dissecting actions performed on structures like the cystic artery, cystic duct, and gallbladder. The model demonstrates strong capability in disentangling overlapping actions, effectively distinguishing interactions involving multiple instruments and anatomical targets within a single frame. In contrast to prior models prone to confusion under such complex scenes, SurgMotion maintains consistent triplet predictions that align with the ground truth annotations. These results indicate that our video-native pretraining enables robust spatiotemporal understanding of fine-grained surgical dynamics and inter-object relationships.

\section{Conclusions}

In this study, we present SurgMotion, a video-native foundation model for universal surgical video understanding. By shifting from pixel-level reconstruction to latent motion prediction through the V-JEPA framework, SurgMotion learns robust spatiotemporal representations that capture the semantic structure of surgical procedures while disregarding low-level visual noise. Our key technical contributions, motion-guided latent masked prediction, spatiotemporal affinity self-distillation, and spatiotemporal feature diversity regularization, specifically address the challenges of surgical videos, including texture-sparse regions, visual artifacts, and the struggle to maintain temporal coherence. To enable large-scale pretraining, we have curated SurgMotion-15M, the largest multi-modal surgical video dataset to date, spanning 3,658 hours across 13 anatomical regions.

Extensive experiments have demonstrated that SurgMotion achieves state-of-the-art performance across a comprehensive variety of surgical video understanding tasks. On workflow recognition, SurgMotion outperforms prior methods on all 8 benchmarks, with particularly strong gains on challenging domains such as EgoSurgery (+14.6\% F1 over the best surgical foundation model) and PitVis (+10.3\% F1). On fine-grained action understanding, SurgMotion sets new state-of-the-art results on triplet recognition, action recognition, and skill assessment. On dense prediction tasks, SurgMotion achieves superior polyp segmentation, especially under domain shift, and best-in-class depth estimation, despite not being explicitly trained for geometric reasoning.

While SurgMotion-15M is the largest surgical video dataset to date, certain procedures and institutions remain underrepresented. Our results demonstrate that strong cross-procedural generalization is achievable when diverse, large-scale data is available, underscoring the importance of continuing multi-institutional data curation efforts.

SurgMotion represents a step toward surgical AI that generalizes across procedures, institutions, and imaging modalities. We envision applications in intraoperative decision support, automated documentation, and objective skill assessment, and hope that SurgMotion will serve as a foundation for further research in surgical video understanding.

\section*{Acknowledgements}
This work was supported in part by the InnoHK Program of the Hong Kong SAR Government and the National Natural Science Foundation of China (Grant No. \#62306313). We also acknowledge the Guangdong Medical Doctor Association Postgraduate Medical Education Innovative Research Project (Project No. \#BJ202602). Felix Holm was supported by a fellowship within the IFI programme of the German Academic Exchange Service (DAAD), funded by the Federal Ministry of Education and Research (BMBF).

\FloatBarrier 
\bibliographystyle{IEEEtran}

\bibliography{ref}

\end{document}